\documentclass[runningheads]{llncs}
\usepackage[table,xcdraw]{xcolor}
\usepackage{times}     
\usepackage{xcolor}
\usepackage{soul}
\usepackage{xcolor}

\usepackage{graphicx}
\usepackage{comment}
\usepackage{amsmath}
\usepackage{float}
\usepackage{soul}
\usepackage{mathtools} 
\DeclarePairedDelimiter{\floor}{\lfloor}{\rfloor}
\DeclarePairedDelimiter{\ceil}{\lceil}{\rceil}
\usepackage{tikz}

\usepackage{ulem}

\begin{document}
\date{}


\title{\Large\textbf{Partition Pruning: Parallelization-Aware Pruning for Deep Neural Networks  
}}
\author{Sina Shahhosseini\and
Ahmad Albaqsami \and
Masoomeh Jasemi++
\and 
Nader Bagherzadeh
}
\authorrunning{S. Shahhosseini, A. Albaqsami, M. Jasemi, N. Bagherzadeh}
\titlerunning{Partition Pruning: Parallelization-Aware Pruning for Deep Neural Networks}
%
\institute{University of California,Irvine \\
\email{\{sshahhos,aalbaqsa,mjasemi,nader\}@uci.com}\\
\email{\ ++Jasemi@ce.sharif.edu}
}
\maketitle
\thispagestyle{empty}

\begin{abstract}
 Parameters of recent neural networks require a huge amount of memory. These parameters are used by neural networks to perform machine learning tasks when processing inputs.
. To speed up inference, we develop Partition Pruning, an innovative scheme to reduce the parameters used while taking into consideration parallelization. We evaluated the performance and energy consumption of parallel inference of partitioned models, which showed a 7.72x speed up of performance and a 2.73x reduction in the energy used for computing pruned layers of TinyVGG16 in comparison to running the unpruned model on a single accelerator. In addition, our method showed a limited reduction some numbers in accuracy while partitioning fully connected layers.

\end{abstract}

\section{Introduction}
Neural networks have become ubiquitous in applications that include computer vision, speech recognition, and natural language processing. The demand for processing neural network applications on edge devices, including smart phones, drones, and autonomous vehicles, is increasing~\cite{13}.  Meanwhile, the size of neural network models has been drastically increased over time, reaching beyond the Peta scale~\cite{13}.
 In 1998, a handwritten digits classifier had about 1 M parameters~\cite{2}, but in 2012, an image classifier for the ImageNet~\cite{12} dataset had more than 60 M parameters. In addition, Neural Talk, which automatically creates proper captions for ImageNet dataset has more 230 M parameters~\cite{18}.
The top 5 error accuracy has been reduced by 30\% each year, suggesting why this trend drastically increases the number of layers, parameters, and operations~\cite{13}.

Large deep neural networks (DNNs) models consume a significant amount of energy because they are required to be stored in DRAMs or on-chip SRAMs, and thus are fetched every time they are processed. From 2012 to 2015, the energy efficiency of DRAMs increased due to CMOS scaling based on Moore's Law. As of 2015, CMOS scaling no longer provided substantial improvements in either energy efficiency or memory density. Because SRAM is realized using CMOS transistors, its energy efficiency is typically bounded by Moore's Law~\cite{sina1}~\cite{sina2}. Therefore, the energy efficiency of the memory cannot keep up with the increasing size of the neural networks. This leads to consuming more energy to accomplish the same processing tasks. Therefore, innovations in architectural design, algorithms development, and circuit technique are required~\cite{15}. 

   Both memory footprint and computational complexity lead to the need for sparsity and/or reducing the number of parameters in a neural network. For example, AlexNet requires 234 MB of memory space for storing parameters and requires 635 million arithmetic operations for feed-forward processing. AlexNet's convolutional layers are locally connected, but they are followed by fully connected layers that make up 95\% of the connections in the AlexNet network~\cite{14}.
Fully connected layers are over-parameterized and tend to overfit the training data. At the algorithm level, pruning methods were proposed before deep learning became popular. Based on the assumption that many parameters are unnecessary, pruning methods remove these parameters, resulting in expanding sparsity of layers~\cite{9}.

	Previous research has sought to reduce the number of parameters. Dropping out random connections was proposed by~\cite{10}. The Optimal Brain Damage~\cite{19} and Optimal Brain Surgeon~\cite{20} reduced the number of connections according to the loss function. Singular value decomposition (SVD) decreased the number of weights~\cite{22}. Another approach , adopted by the GoogleNet model~\cite{21}, exploits the convolutional layers rather than the fully connected layers. This resulted in sparse layers that provided three benefits~\cite{17}. First, sparse layers required less storage for space for parameters. Second, it omitted computation of the removed edges, which reduced power consumption and latency. Third, it required less memory bandwidth to transfer parameters from the DRAM.      
    
	 In this paper based on the insight that smart pruning can reduce the number of off-chip accesses, we propose a new scheme to better partition and prune the inputs to each layer. This way, we partition a large matrix into small matrices and distribute them to multiple computational units. The proposed partitioning algorithm has three objectives: first enhancing the parallelism among accelerators, second reducing the number of off-chip accesses, and third maintaining the accuracy as high as the baseline. In the first step, we formulate the problem and then enforce some constraints. We define our constraint in such a way that the three mentioned objectives are satisfied.  .The experimental results show that the proposed scheme can increase the speed up by 7.72x and energy efficiency by 2.73x, respectively.

The rest of this paper is organized as follows. Section. \ref{section:overview} provide an overview of the problem. Section \ref{sectio:PP} describes the proposed partition pruning algorithm followed by Multi-core organization in Section. \ref{section:Multicore}. Experimental setup and evaluation methodology is presented in Section. \ref{section:evalaution}. We discuss the result in Section. \ref{section:results}. And finally, we conclude the paper in Section. \ref{section:conclusion}.

\section{Overview}\color{black}
\label{section:overview}

\begin{figure*}[h] 
	\centering
		\includegraphics[width=1 \textwidth]{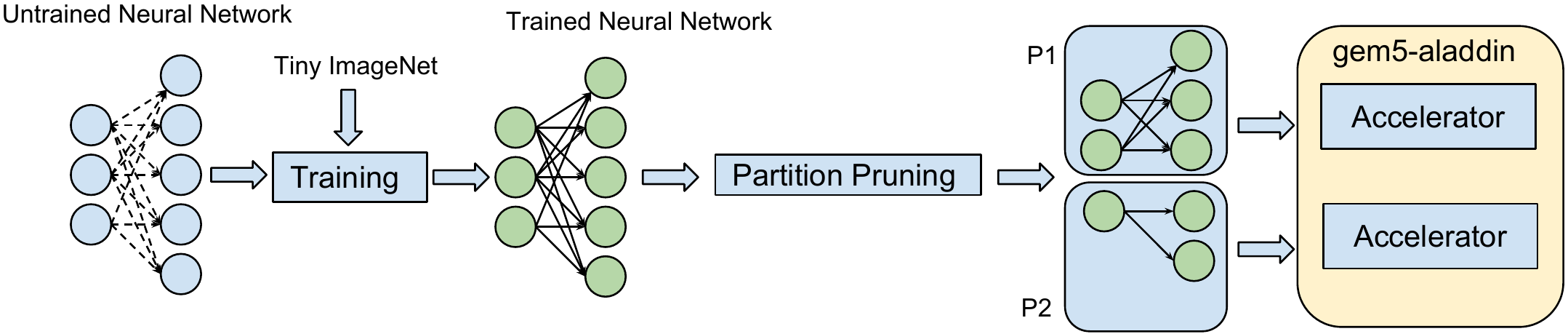}
	\caption{Overview of the procedure used. Note that Partition Pruning is applied to a trained neural network since it is dependent on the weights of the fully connected layer(s). The illustration shows only one fully connected layer.}
	\label{fig:overview}
    \vspace{-13.5pt}
\end{figure*}
Figure~\ref{fig:overview} illustrates a high-level diagram of the proposed framework. First, a neural network model is trained. Section V discusses the baseline accuracy for different neural network models that were used to evaluate the framework. Then, fully connected layers of each model were pruned using the Partition Pruning approach. Section III explains how the partitioning algorithm was applied to these layers. Then, inference was performed on multiple processing cores. Section IV explains multi-core architecture, which provides the ability to run parallel matrix multiplication. Section VI evaluates our framework in terms of performance and accuracy.\color{black}

\section{Partition Pruning}
\label{sectio:PP}

\subsection{System Model}
\begin{figure} [!tb]
	\centering
		\includegraphics[width=0.75 \textwidth]{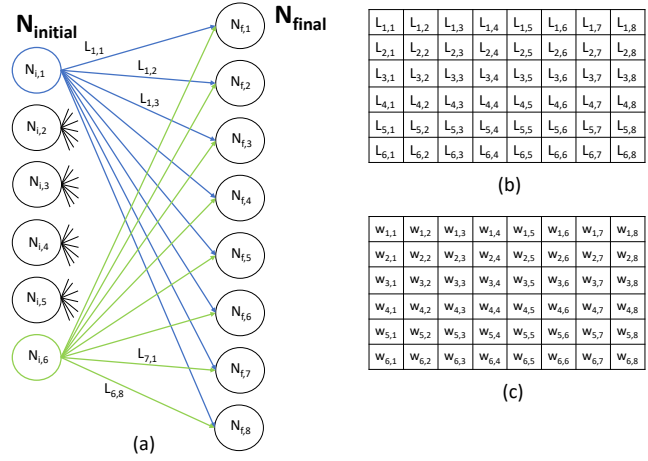}
	\caption{Example of a model representation of a fully connected layer. b) shows the connection's representation in matrix form. Note that in the above case,$C=C_{full}=6\times 8=48$ and $R=1$. }
	\label{fig:model}
    \vspace{-13.5pt}
\end{figure}

Our framework targets neural networks that have some or all of their nodes fully connected to the subsequent nodes. The \textit{set} of starting nodes, \boldmath{$N_{initial}$} is \textit{fully} connected to the subsequent nodes \boldmath{$N_{final}$},  i.e. \textit{fully-connected layers}. A link, which is a parameter, is a connection represented by $L_{ij}$, where $i$ is the starting node number and, $j$ is the connected node number within a layer. The link's value (i.e the parameter's weight) is represented by $w_{i,j}$. $L_{i,j}=0$ if the link is pruned, and if not, $L_{i,j}=1$. Note that $w_{i,j}$ may contain any value. The set of weights, \boldmath{$W_i$}, consists of links, \boldmath{$L_i$}, that connect between the set of Nodes, \boldmath{$N_{i}$}, and \boldmath{$N_{j}$}. Figure~\ref{fig:model}a shows an example of a fully connected layer of size $6\times 8$. Figure~\ref{fig:model}b shows the matrix representation of the fully connected layer. While Figure~\ref{fig:model}c indicates the weight matrix of the fully connected layer.
The \textit{connectedness number}, $C$, is simply;
\begin{equation} \label{eq:conectedness}
C=\sum_{i=1}^{|N_{initial}|} \sum_{j=1}^{|N_{final}|} L_{i,j}
\end{equation}
\color{black} A fully connected layer is annotated as $C_{full}$ and thus;
\begin{equation} \label{eq:full}
C_{full}=|N_{initial}| \times |N_{final}|
\end{equation}
Therefore, the \textit{connectedness ratio}, $R$, is:
\begin{equation} \label{eq:ratio}
R=\frac{C}{C_{full}}
\end{equation}

\begin{figure} [!tb]
	\centering
		\includegraphics[width=0.75 \textwidth]{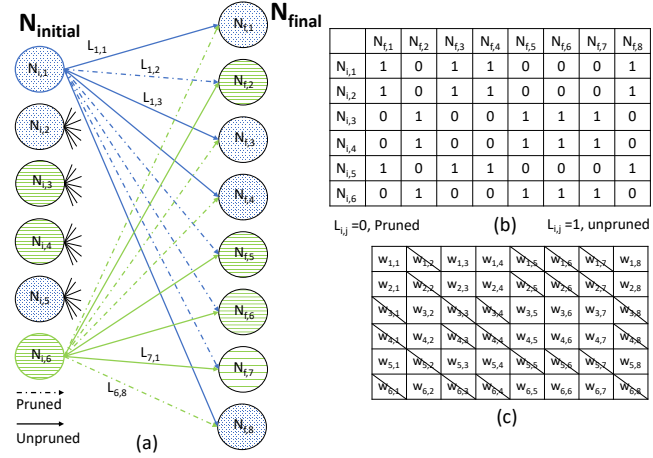}
	 \caption {a) indicating what links are to be pruned from the fully connected layer  b) shows the connection's representation, with 0s representing the absence of a link. Note that in the above case,$C=C_{full}=12$ and $R=0.5$.}
	\label{fig:model_ex2}
    \vspace{-13.5pt}
\end{figure}

\begin{figure} [!tb]
	\centering
		\includegraphics[width=0.75 \textwidth]{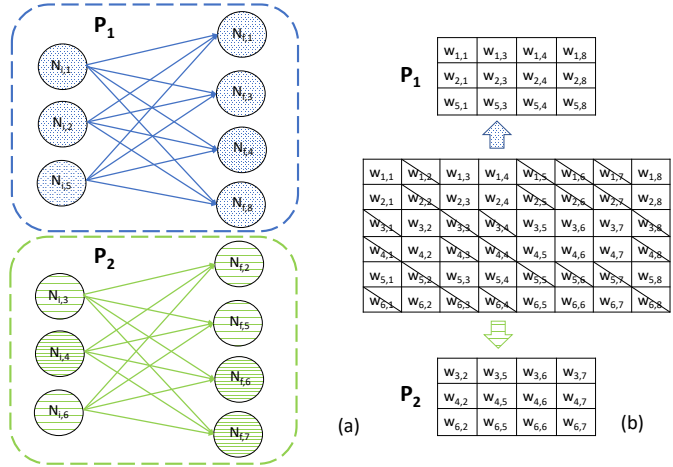}
	\caption{a) the resulting partitions shows full independence. b) shows the resulting reduction of parameters due to the 2-partition targeted pruning.}
	\label{fig:twopart}
    \vspace{-13.5pt}
\end{figure}

Figure~\ref{fig:model_ex2} shows an example of a 2-partition pruning of the fully connected layer from Figure~\ref{fig:model}. Figure~\ref{fig:twopart} visually illustrates the partitions of Fig~\ref{fig:model_ex2} and the reduction of number of weights due to that partitioning. Given that there are $|P|$ partitions, where a $P_x\in \boldmath{P}$, then any given $N_{initial,j} \in P_x$ will not be in any other partition. The same goes for nodes in $N_{final,i}$ . More formally,
\begin{equation} \label{eq:constraint}
\{P_i,P_j \in \boldmath{P}|i\neq j , P_i \cap P_j = \emptyset \}
\end{equation}
Equation \ref{eq:constraint} is the constraint of the groupings of nodes in $\boldmath{N_{initial}}$ and $\boldmath{N_{final}}$. That is, once a particular node is in a particular partition, it cannot be a member of another partition. Another way of stating this is:
\begin{equation} \label{eq:nodeconstraint}
N_{i}\in P_n \text{ Then } N_i \not\in P_m,\forall m \neq n
\end{equation}
Note that there is an {upper,$\ceil*{\frac{|N_{initial}|}{|P|}}$, and lower,$\floor*{\frac{|N_{initial}|}{|P|}}$, bound} to the number of $N_{initial,i}$ nodes that are members of a partition $P_n$. The same is true for $N_{final,i}$ nodes. In addition, the number of partitions that contain the upper limit is $|N_{initial}| \mod |P|$, while the number that contain the lower limit is $|P| - (|N_{initial}| \mod |P|)$.\color{black} As an example, if $|N_{initial}|=22$ and $|P|=5$ (i.e number of partitions), then an example of partition sizes for $N_{initial}$, ignoring $N_{final}$, would be $$(|P_1|,|P_2|,|P_3|,|P_4|,|P_5|)=(4,5,4,4,5)$$ 
Therefore, the example suggests that there are three partitions of size 4 and two partitions of size 5. This bound description also applies to $N_{final}$. 
\subsection{Partition Pruning Overview}
The objective of Partition Pruning is two-fold: pruning with the objective of having balanced partitions, and pruning with the objective of having the least absolute weight-loss. The second objective guarantees a smaller loss of accuracy, while the first allows for maximum parallelism. 
Note that the number of parameters pruned is directly related to the number of partitions desired. The connectedness ratio, in relation to the number of partitions is $R_{|P|} = \frac{1}{|P|}$.
Thus, for a given $|P|$, Partition Pruning will find the following:
\begin{equation*} \label{eq:objective}
\begin{aligned}
& \underset{x}{\text{min}} 
& &|C_{full}\sum{|w_{i,j}}| - \sum{x_{i,j}|w_{i,j}}|| \\
& \text{subject to}
& &  x_{i,j} = 0 \text{ or } 1\\
&&&\sum{x_{i,j}=R_{|P|}C_{full}}\\
&&& \{P_m,P_n \in \boldmath{P}|n \neq m , P_m \cap P_n = \emptyset \}
\end{aligned}
\end{equation*}
From the objective function, we determine which $1-R_{|P|}C_{full}$ parameters are pruned for a particular fully connected layer while minimizing the cumulative weight-loss.

\subsection{Input/Output}
The input to the Partition Pruning algorithm is a matrix representation, $W_{fc,i}$, of the targeted fully connected layer, $i$. This is exemplified in Figure~\ref{fig:model}c. Note that the fully connected layer is assumed and asserted to be trained. That is, the parameters have the correct values for the targeted neural network's base accuracy. In a fully connected layer, every element of the matrix $L_{fc,i}$ is 1 (see Equation \ref{eq:full}). After Partition Pruning, the output will be $L_{part,i}$ and the sum of all its elements would be $RC_{full}$. This is exemplified in Figure~\ref{fig:model_ex2}b.

\color{black}\subsection{Methodology}
This section  the methodology of selecting the links to prune, taking into consideration the partitioning. The example of $|N_{initial}|=7$, $|N_{final}|=10$, and $|P|=3$, will be used to describe the process. Figure~\ref{fig:overview} \color{black}shows an overview of the methodology and where Partition Pruning resides.
\newline


\textbf{ Start: Selection of $N_{initial,i}$, and $N_{final,j_1,j_2..}$:}

\begin{figure} [!tb]
	\centering
		\includegraphics[width=0.75 \textwidth]{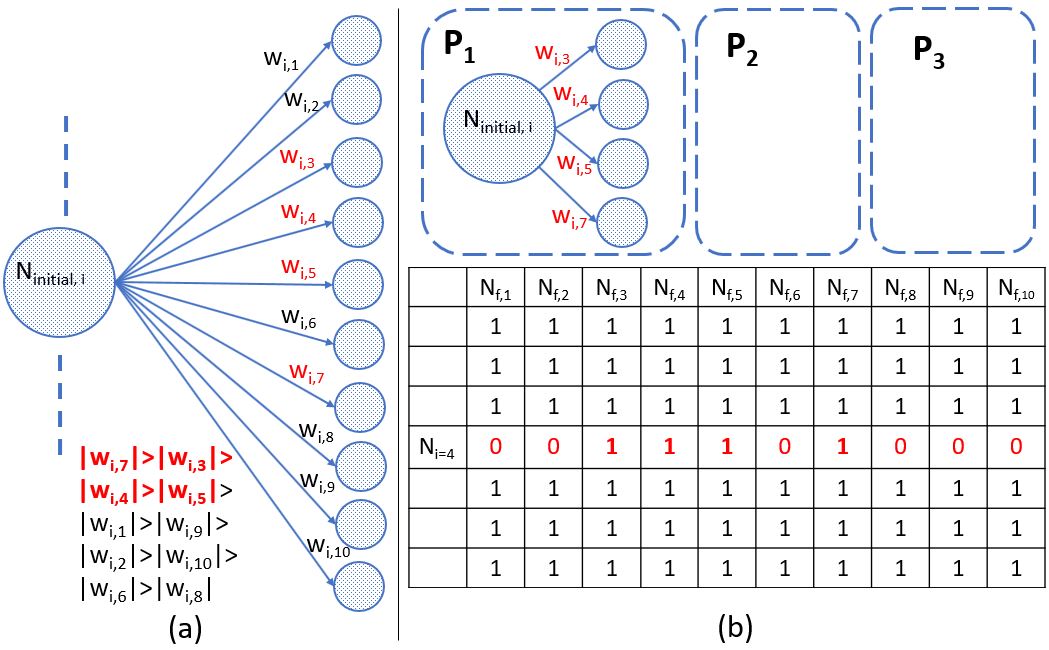}
	\caption{Random selection of $N_{initial,i}$, where $i=4$ in this example. The top four weights, in terms of magnitude, are $w_{i,7},w_{i,3},w_{i,4}$, and $w_{i,5}$ in descending order. Note that its top \textit{four} because of the upper bound, $\ceil*{|N_{final}|/|P|}=\ceil*{10/3}=4$ b) $P_1$, after partitioning, contains four nodes (the limit) from $N_{final}$, and one node from $N_{initial}$.The $L$ matrix is updated for row i=4}
	\label{fig:select1}
    \vspace{-13.5pt}
\end{figure}

In the first stage, a row in the matrix is \textit{randomly} selected. That is, a random $N_{initial,i}$ is selected for processing. Note that currently $|P_n|=0$ for all $n$, because no pair of nodes, \color{black} has joined a partition. After choosing an $N_{initial,i}$, a set of $N_{final}$ nodes is chosen, and in this case, the set size is $\ceil*{\frac{|N_{final}|}{|P|}}$. The node $N_{initial,i}$, and the nodes $N_{fianl,j_1,j_2..}$ are chosen to be part of the first partition, $P_1$. Those selected will have their $L_{i,j}=1$, while those not selected will have their $L_{i,j'}=0$. Note that the links selected have the \textit{highest magnitudes} (refer to Figure~\ref{fig:select1}a as an example). Figure~\ref{fig:select1}b illustrates an example of the change in values and a pictorial representation of the first partition.
\newline

\textbf{ Non-Start: Selection:}

Moving forward, another $N_{initial,i}$ node is selected at random. The highest, non-partition members, $w_{i,j}$s are sorted from the highest to the lowest magnitude, as was done previously. The sum of the highest upper bound (or a lower bound if all upper bound partitions are fulfilled) are compared with the sum of the magnitude of partition-member weights/links that still have capacity (as per the upper and lower bounds of the number of nodes of type $N_{initial}$). 

\begin{figure} [!tb]
	\centering
		\includegraphics[width=0.75 \textwidth]{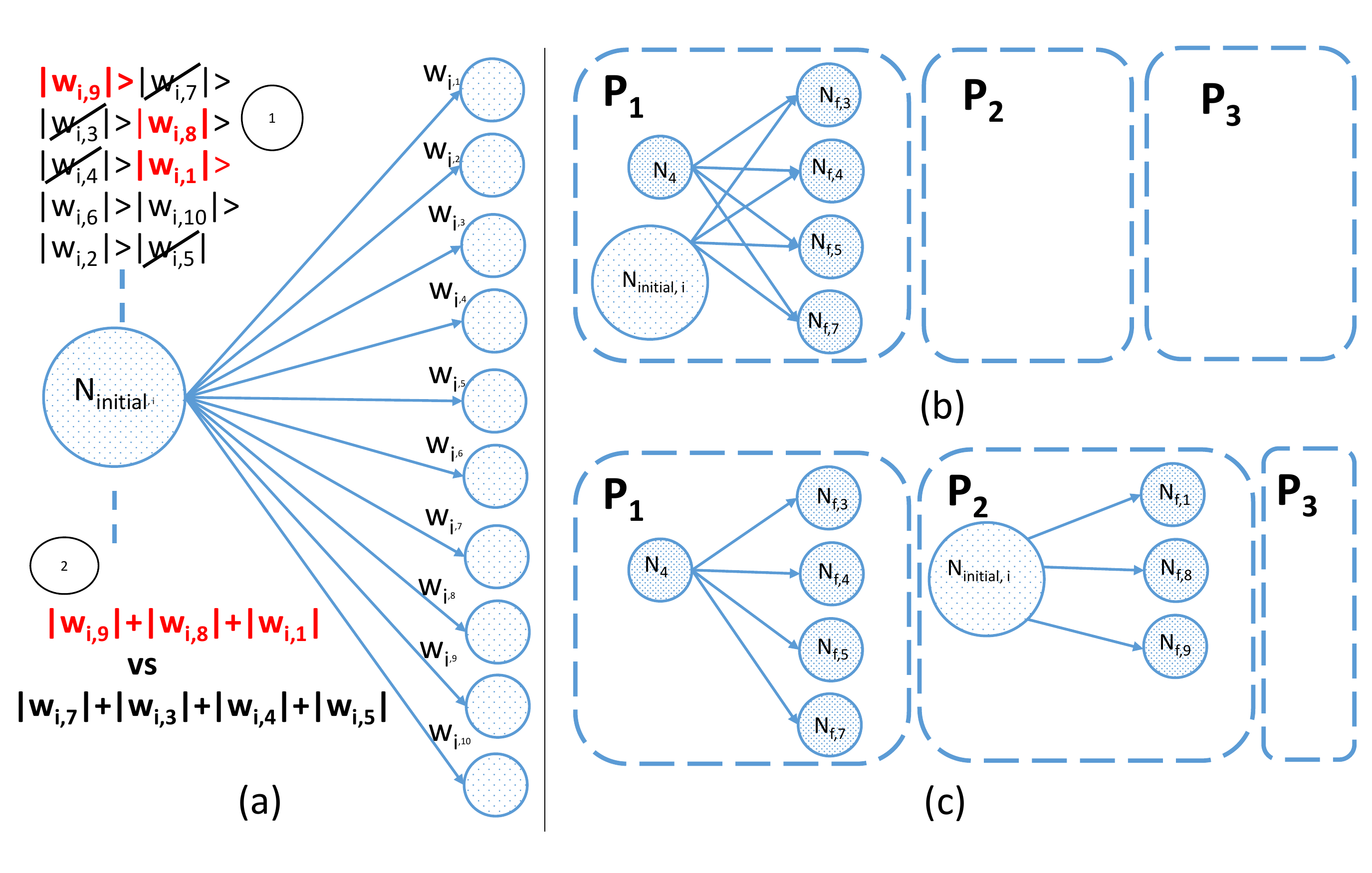}
	\caption{second random selection of $N_{initial,i}$ (where $i \neq 4)$. The top three weights (1), in terms of magnitude and are none partition members, are $w_{i,9},w_{i,8},w_{i,1}$, in descending order. Note that its top \textit{three} due to the the capacity for $N_{final}$ node type is $(P_1,P_2,P_3)=(4,3,3)$ b) shows the situation in case of $|w_{i,7}|+|w_{i,3}|+|w_{i,4}|+|w_{i,5}|>|w_{i,9}|+|w_{i,8}|+|w_{i,1}|$. c) is the case scenario.}
	\label{fig:select2}
    \vspace{-13.5pt}
\end{figure}

\textbf{ End and Try Again:}

This process is repeated until every partition $P_m$, is at capacity in terms of $N_{initial}$ nodes and $N_{final}$ nodes. Note that the partitioning is dependent on which row, i.e $N_{initial,i}$ was selected at each iteration. Once the process is completed, the weight-loss is recorded. 

\section{Multi-Core Organization}
\label{section:Multicore}

\begin{figure} [!tb]
	\centering
		\includegraphics[width=0.75 \textwidth]{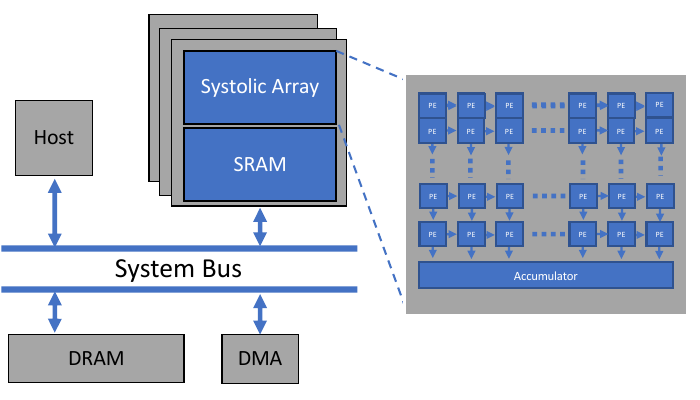}
	\caption{ Architectural template for generated accelerators.}
	\label{fig:arch}
    \vspace{-13.5pt}
\end{figure}

Figure \ref{fig:arch} shows the architecture of an System on Chip (SoC) that consists of general purpose cores, memory controllers, a DMA engine, and matrix multiplication accelerators all of which are connected through the system bus. To understand how the system level affects the accelerators' behavior, simulation infrastructures that can model these heterogeneous systems are needed. gem5-Aladdin system simulator is used to evaluate the proposed architecture. This tool is an integration of a gem5 system simulator with an Aladdin accelerator simulator. It is a pre-RTL simulation infrastructure that models multiple accelerators and interactions with central processing units (CPUs) in an SoC that consists of Processing Elements (PEs), fixed-function accelerators, memory controllers, and interfaces. This simulator can model the accelerators' performance, area, and power~\cite{gem5-aladdin}\cite{aladdin}. Multiple matrix multiplication units are connected to the bus. In the gem5-Aladdin system, the accelerators can invoke the DMA engine already present in the Gem5. The DMA is used to transfer bulk data without the CPU's intervention. The internal SRAM stores the weights, input features, and the outputs of the matrix multiplication. Each accelerator uses a 32 x 32 Systolic Array (SA). The SA architecture is a specialized form of parallel computing in which tightly coupled processing elements are connected to a small number of their nearest neighbors in a mesh-like topology. This architecture has a very low amount of global data transfer and can achieve a high clock frequency. However, SA architecture suffers from scalability issues due to the shape being fixed. 

In an SA, the horizontal systolic movements are for implementing data broadcasts, and the vertical ones are for implementing accumulations.

\begin{table}[!tb]
\centering
\caption{System Configuration Parameters}
\begin{tabular}{|c|c|}
\hline
{ Parameter} & { Value} \\ \hline
Host Clock Frequency             & 1 GHz                        \\ \hline
Accelerator Clock Frequency      & 200 MHz                      \\ \hline
Technology Width                 & 40 nm                        \\ \hline
DRAM                             & DDR3-1600-8x8                \\ \hline
Number of CPU                    & 1                            \\ \hline
Systolic Array Size              & 32x32                        \\ \hline
Data Type                        & FP-32                        \\ \hline
Data Transfer                    & DMA                          \\ \hline
\end{tabular}

\end{table}

\section{Experimental Setup}
\label{section:evalaution}
Fully connected layers are pruned by using Partition Pruning for three networks that use a TinyImageNet~\cite{23} dataset. which consists of 100,000 training images, 10,000 validation images, and 10,000 testing images that have dimensions of 64x64x3, and that classify 200 labels. These images are taken from the ImageNet~\cite{12} dataset, cropped into squares, and resized to 64x64. For each network, the fully connected layers are partitioned to 2, 3, 4, and 5 partitions, resulting in the pruning of 50\%, 66\%, 75\%, and 80\%, of the fully connected links, respectively. 
	
    Initially, the neural networks are trained and evaluated on a TinyImageNet dataset, as shown in Table \ref{tab:benchmarks}. 
Convolutional neural networks represent the state-of-the-art in image classification. AlexNet~\cite{7} and VGG16~\cite{2} are well-known deep convolutional neural networks that have previously won ImageNet competitions. TinyVGG16 and TinyAlexNet use a 56x56x3 input image instead of 228x228x3, as do the original VGG16 and AlexNet. Each network has three fully connected layers at the end its structure. Partition Pruning prunes the first two of these three fully connected layers. The omission of pruning the last fully connected layer is due to the fact that every link is required for classification. If pruned, the classification accuracy would be affected the considerably and detrimental to the performance of the Neural Network model. Table \ref{tab:benchmarks} shows the benchmarks' baseline performances. 
After training the networks, Partition Pruning is applied to two, of the three, fully connected layers. Google's TensorFlow~\cite{6} version 1.7 was used to model the benchmarks. Partition Pruning was implemented in Python 2.7 and was given the NumPy matrices from the first two fully connected layers of the benchmarks. Then, the weights were updated in the TensorFlow model files using the resulting output filters. Note that, as mentioned earlier, gem5-Aladdin is used to evaluate the performance.

\begin{table}[tbp]
\centering
\caption{Baseline Top-5 and Top-1 accuracy for VGG16, AlexNET}
\begin{tabular}{|c|c|c|}
\hline
{Network Name} & {Top-5 Accuracy} & {Top-1 Accuracy} \\ \hline
{\color[HTML]{333333} TinyVGG16}    & {\color[HTML]{333333} 76.96\%}        & {\color[HTML]{333333} 52.41\%}        \\ \hline
{\color[HTML]{333333} TinyAlexNet}  & {\color[HTML]{333333} 72.06\%}        & {\color[HTML]{333333} 46.73\%}        \\ \hline
\end{tabular}

\label{tab:benchmarks}
\end{table}

\section{Results}
\label{section:results}
\begin{figure} [!tb]
	\centering
		\includegraphics[width=1 \textwidth]{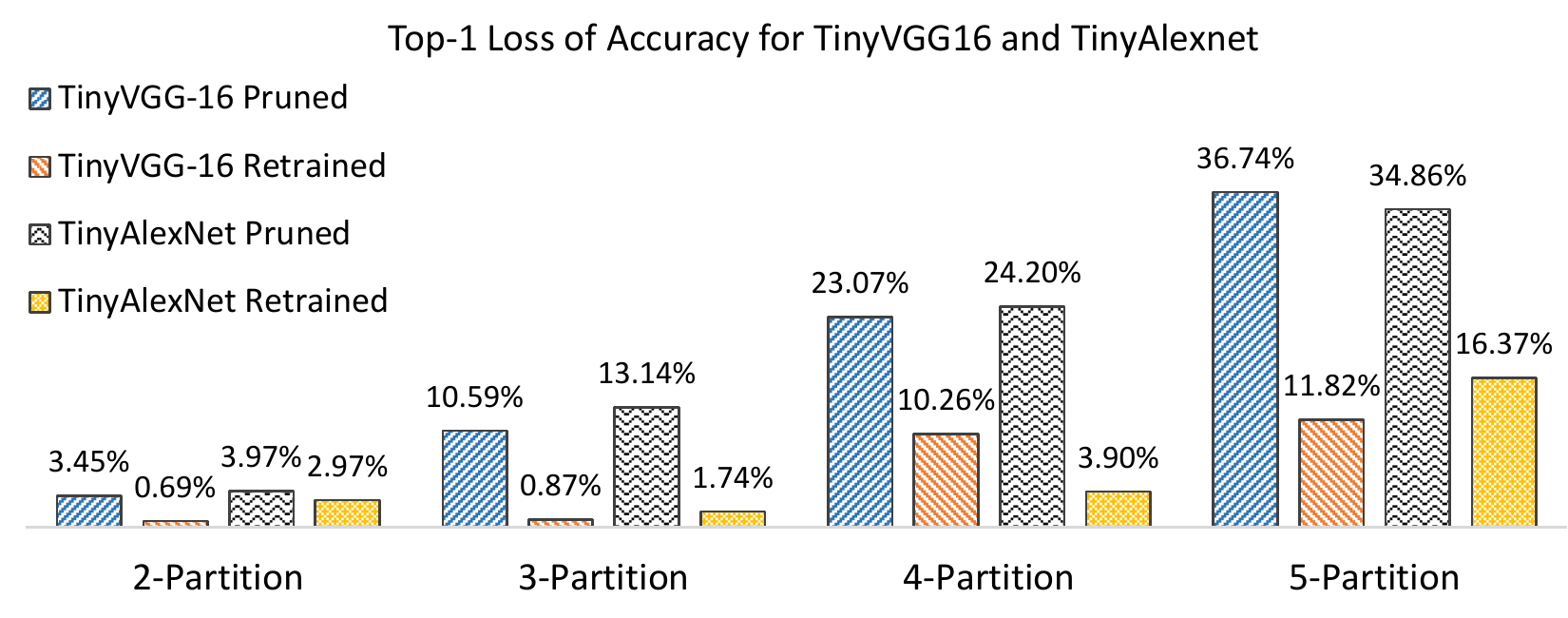}
	\caption{Top-1 loss of Accuracy for VGG16 and Alexnet. Note that the number of links pruned is equal within the partition group. In retraining, only the
non-pruned links are retrained.}
	\label{fig:accloss}
    \vspace{-13.5pt}
\end{figure}

\begin{figure} [!tb]
	\centering
		\includegraphics[width=0.75 \textwidth]{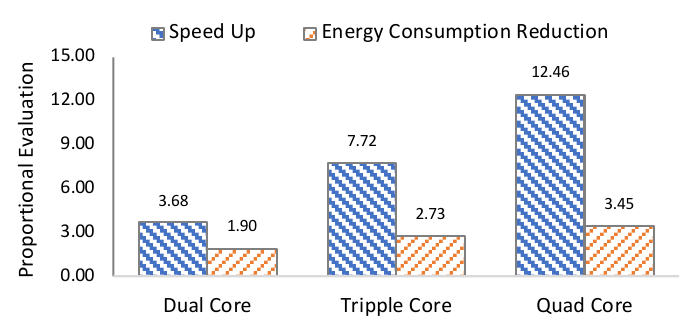}
	\caption{Power and performance of multi-accelerators are shown. Results are evaluated proportional to result of unpruned benchmark on single core accelerator. }
	\label{fig:powerperf}
    \vspace{-13.5pt}
\end{figure}
Table \ref{tab:benchmarks} shows the initial baseline accuracies, without pruning, of the TensorFlow implementations of the neural network benchmarks. Figure \ref{fig:accloss} shows the resulting accuracy losses of the Partition Pruning algorithm for TinyVGG16 and TinyAlexNet. Note that results for retraining are also shown. Accuracy loss increases when the number of partitions is increased, given that more parameters are pruned. After pruning, retraining the models reduces the loss of accuracy. For example, in 3-Partition, retraining reduces accuracy loss in TinyVGG16 from 10.59\% to 0.87\%. As Figure \ref{fig:arch} shows, running inference of partitioned TinyVGG16 layers on different accelerators speeds performance and reduces energy consumption. These results are in comparison to running inference of the unpruned layers on signle accelerator. For example, running this benchmark on a triple-core accelerator executes 7.72x faster while consuming 2.73x less energy. This is because pruning reduces the size of the benchmarks by a factor correlated to the partition number (for example, by a factor of 2x for two partitions). In addition, running inference in parallel on multiple accelerators speeds the execution time. Therefore, the performance speed and the energy consumed by processing partitioned models were both improved by reducing the size of the models and using multiple hardware resources. Running the same benchmarks on multiple accelerators does not increase speed as expected. For example, running two identical workloads on two accelerators can increase speed 1.8x, and on three accelerator, 2.5x. This happens because all accelerators are connected to the same bus with one DMA, which leads to bus congestion. It is expected that using multiple large SAs, for example 256 x 256, would cause bandwidth bottlenecks and sizeable bus congestion. Although using a small SA does not provide high throughput processing, it leads to low power design because of the number of processing elements used in each accelerator.

\section{Conclusions}
\label{section:conclusion}

This paper presented Partition Pruning, an approach that prunes fully connected layers of neural network models with the aim of partitioning for parallelization in order to improve speed and energy. The idea behind Partition Pruning approach is to target low overall weight loss to reduce the impact on accuracy. The approach shows that by partitioning fully dense layers of TinyVGG16 to 3-Partition and executing the model on multiple accelerators, a speed increase of 7.72x and an energy reduction of 2.73x can be obtained. Future work will evaluate a system that has multiple high-bandwidth memories and neural network accelerators. In addition, more optimizations will be applied to the accelerators to minimize power consumption and increase throughput.

\end{document}